\documentclass[10pt,journal,cspaper,compsoc]{IEEEtran}
\usepackage{}

\ifCLASSOPTIONcompsoc
\else
\fi

\ifCLASSINFOpdf
   \usepackage[pdftex]{graphicx}
   \graphicspath{{../pdf/}{../jpeg/}{../png/}}
   \DeclareGraphicsExtensions{.pdf,.jpeg,.png}
\else
   \usepackage[dvips]{graphicx}
   \graphicspath{{../eps/}}
   \DeclareGraphicsExtensions{.eps,.png}
\fi

\ifCLASSOPTIONcompsoc
\usepackage[tight,normalsize,sf,SF]{subfigure}
\else
\usepackage[tight,footnotesize]{subfigure}
\fi

\begin{document}

\title{A novel automatic thresholding segmentation method with local adaptive thresholds}

\author{Bo~Xiao,~
        Yuefeng~Jing,~
        and~Yonghong~Guan
\IEEEcompsocitemizethanks{\IEEEcompsocthanksitem B. Xiao, Y. Jing, and Y. Guan
are with the Institute of Fluid Physics, China Academy of Engineering Physics.\protect\\
E-mails: homenature@caep.ac.cn, jyf@caep.ac.cn, yhg@caep.ac.cn, respectively
}
\thanks{}}


\IEEEcompsoctitleabstractindextext{%
\begin{abstract}

A novel method for segmenting bright objects from dark background for grayscale image is proposed.
The concept of this method can be stated simply as: to pick out the local-thinnest bands on the grayscale grade-map.
It turns out to be a threshold-based method with local adaptive thresholds, where each local
threshold is determined by requiring the average normal-direction gradient on the object boundary to
be local minimal. The method is highly
automatic and the segmentation mimics a man's natural expectation even the object boundaries are fuzzy.

\end{abstract}

}

\maketitle

\IEEEdisplaynotcompsoctitleabstractindextext

\IEEEpeerreviewmaketitle

\section{Introduction}

\IEEEPARstart{O}{bject} edge extraction (segmentation) is one of the oldest yet
still the most fundamental task in vision and image analysis \cite{Chan:2005}.
In the study of the computer aided detection/diagnose (CAD) of medical images, e.g. mammography or breast ultrasound,
the mass segmentation is an important basic task, whose accuracy determines the accuracy of the final diagnose \cite{McClymont:2012}.
Active contour is one of the most frequently used methods (e.g. \cite{Huang:2007,Oliver:2010,Chang:2011}).
However, because active contour method needs an initial contour,
authors have to part the active contour with other preprocessing techniques, such as
thresholding segmentation \cite{Huang:2007}, to achieve full automatic segmentation, as required by automatic CAD systems.

Considering it's a simple and fast procedure for a man to pick out masses in an image,
we believe that simple and fast segmentation methods should be there.
In this paper we study a new segmentation method, which implements the
grayscale and boundary gradient information (what we believe to be the main information used
in a man's fast bare-eye recognition) in a simple way.
This method realizes, hopefully, the vision.

The remainder of the article is arranged as follows.
In section \ref{sec.method} the proposed method are described.
In section \ref{sec.test} the method are implemented in different types of images to test its properties.
Conclusions and discussions are given in section \ref{sec.end}.



\section{The new segmentation method}
\label{sec.method}

\begin{figure}[!t]
\centering
\subfigure[]{\includegraphics[width=0.75in]{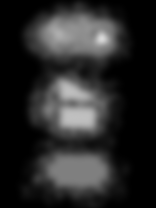}\label{fig_steps_a}}
\subfigure[]{\includegraphics[width=0.75in]{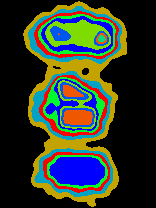}\label{fig_steps_b}}
\subfigure[]{\includegraphics[width=0.75in]{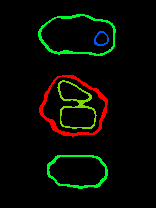}\label{fig_steps_c}}
\subfigure[]{\includegraphics[width=0.75in]{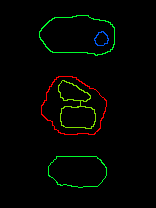}\label{fig_steps_d}}
\caption{Illustrating the basic steps of the new segmentation method.
         (a) is the origin grayscale image (synthetic); (b) is the grayscale grade-map of the image;
         (c) is the LMW bands from the grayscale grade-map;
         (d) is the single pixel boundary got from shrinking the LMW bands.}
\label{fig_steps}
\end{figure}

The basic concept of the new segmentation method is simple:
to pick out each local-thinnest band in the grayscale grade-map. The method goes in three steps,
as illustrated in Fig. \ref{fig_steps}: first, draw the grayscale grade-map of the original grayscale image, Fig. \ref{fig_steps_b}; second,
pick out the local-minimal-wide (LMW) bands on the grayscale grade-map, Fig. \ref{fig_steps_c}; third, shrink the bands to single pixel wide boundary lines, Fig. \ref{fig_steps_d}.

Now to put each step in exact mathematical definition:

1) Suppose that the grayscale of an image ranges from $g_{\rm min}$ to $g_{\rm max}$.
  An iso-grayscale map is drawn according to the $(N+1)$ evenly interpolated grayscale values
  $g_{\rm min}$, $g_{\rm min}+\Delta$, ..., $g_{\rm min}+(N-1)\Delta$, $g_{\rm max}$.
  In the iso-grayscale map, an iso-grayscale line
  $g_{\rm min}+n\Delta$ can enclose several iso-grayscale lines $g_{\rm min}+(n-1)\Delta$.
  The area between the grayscale line $g_{\rm min}+n\Delta$ and the grayscale lines $g_{\rm min}+(n-1)\Delta$
  is called a band. That means, the pixels in a $n$th band are belong to the same grayscale grade
  $(g_{\rm min}+(n-1)\Delta,g_{\rm min}+n\Delta)$, and there can be several separate $n$th bands. Thus at last we get a map composed of bands,
  called grayscale grade-map, as shown in Fig. \ref{fig_steps_b}.

2) Next we need to define the band width and the neighborhood relations of the bands, in order to define the ``local minimal width''. 

We propose the following definition of band width
\begin{equation}
\label{eqn_widthDef}
W_{\rm B}\equiv 2\times(n_{\rm B}) / (n_{\rm E}),
\end{equation}
where $n_{\rm B}$ is the total number of pixels in a band, and $n_{\rm E}$ is the total number
of pixels at both the inner and outer edge of the band (i.e., the pixels on the inner and outer
iso-grayscale lines enclosing the band). The factor 2 is correctly assigned, so that
 the band width $W_{\rm B}$ would equal $(r_2-r_1)$ in the case of a perfect circ band.

\begin{figure}[!t]
\centering
\subfigure[]{\includegraphics[width=0.8in]{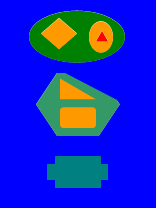}\label{fig_tree_a}}
\subfigure[]{\includegraphics[width=2.2in]{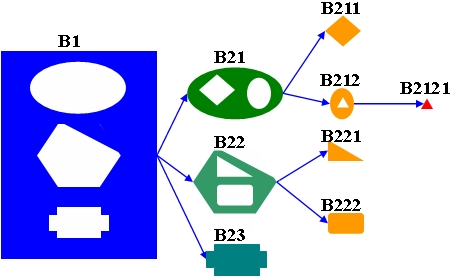}\label{fig_tree_b}}
\caption{The relations of the bands on a grayscale grade-map can be well described by a tree structure.
(a) is a grayscale grade-map (it is a phantom image mimicking the image in Fig. \ref{fig_steps_b});
(b) is the tree structure describing the relations of the bands;}
\label{fig_tree}
\end{figure}

The neighborhood relations of all the bands on the grayscale grade-map can be well described by a tree structure
(which is also the data structure we used in the C-language programming of the method), as
shown in Fig. \ref{fig_tree}. Each band has a ``Father'' neighbor and one or several ``Son'' neighbors,
except for the ``Root'' (which has only Son neighbors) and the ``Leaf''s (which each has only a Father neighbor).
For example, B21 is neighborhood to B2, B211 and B212 in Fig. \ref{fig_tree_b}.

Now we can define a band ${\rm B}_i$ to be a LMW band if
\begin{equation}
\label{eqn_LMWdefine}
W_{{\rm B}_i}\le W_{{\rm B}_k}, \forall {\rm B}_k\in \{{\rm B}_i\texttt{ neighbors}\}
\end{equation}
For example, in Fig. \ref{fig_tree}, the nodes B212 and B22 are LMW bands.

3) Finally, each LMW band is shrunk to a single pixel wide boundary line.
The shrink can be done, for example, by corrosion.

For the above discussion, there is one thing worth noting:
a band can have zero-width in some parts of it (or in its whole body), where the band
disappears on the grayscale grade-map. In fact, those are
the places where the band's outer edge line (the iso-grayscale line $g_{\rm min}+n\Delta$)
coincide with its inner edge lines (the iso-grayscale lines $g_{\rm min}+(n-1)\Delta$),
and thus those places correspond to ideal edges. Those special cases can be treated uniformly with the usual cases
in the computer programming.

\section{Test the method}
\label{sec.test}

\begin{figure}[!t]
\centering
\subfigure[]{\includegraphics[width=1.5in]{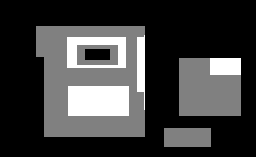}\label{fig_Artif_a}}
\subfigure[]{\includegraphics[width=1.5in]{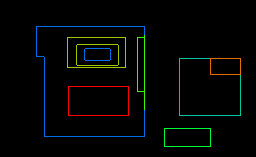}\label{fig_Artif_b}}
\caption{Test of the method on an synthetic image (a). (b) is the segmentation result by our method}
\label{fig_Artif}
\end{figure}

\begin{figure}[!t]
\centering
\subfigure[]{\includegraphics[width=1.0in]{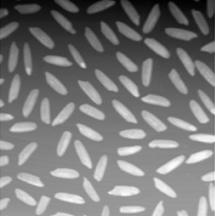}\label{fig_Grain_a}}
\subfigure[]{\includegraphics[width=1.0in]{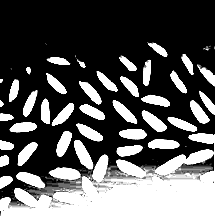}\label{fig_Grain_b}}
\subfigure[]{\includegraphics[width=1.0in]{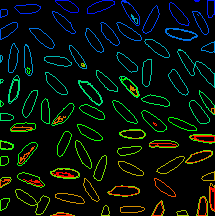}\label{fig_Grain_c}}
\caption{Test of the method on an inhomogeneous illuminated grains image (a). (b) is the segmentation
result by simple hand-tuned thresholding. (c) is the segmentation result by our method.}
\label{fig_Grain}
\end{figure}

\begin{figure}[!t]
\centering
\subfigure[]{\includegraphics[width=1.0in]{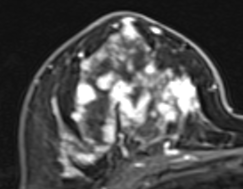}\label{fig_MRI_a}}
\subfigure[]{\includegraphics[width=1.0in]{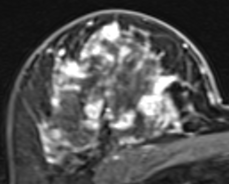}\label{fig_MRI_b}}
\subfigure[]{\includegraphics[width=1.0in]{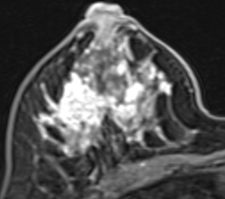}\label{fig_MRI_c}}\\
\subfigure[]{\includegraphics[width=1.0in]{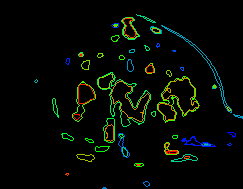}\label{fig_MRI_d}}
\subfigure[]{\includegraphics[width=1.0in]{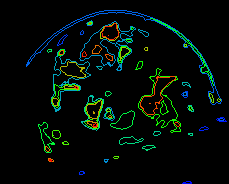}\label{fig_MRI_e}}
\subfigure[]{\includegraphics[width=1.0in]{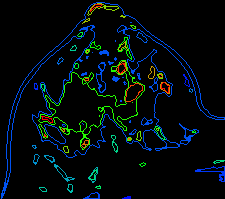}\label{fig_MRI_f}}
\caption{Test of the method on the real breast MRI images. (a)-(c) are the real MRI images.
(d)-(f) are the segmentation results, respectively.}
\label{fig_MRI}
\end{figure}

\begin{figure}[!t]
\centering
\subfigure[]{\includegraphics[width=0.6in]{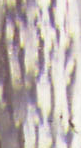}\label{fig_cracks_a}}
\subfigure[]{\includegraphics[width=0.6in]{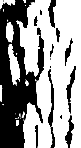}\label{fig_cracks_b}}
\subfigure[]{\includegraphics[width=0.6in]{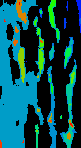}\label{fig_cracks_c}}
\subfigure[]{\includegraphics[width=0.6in]{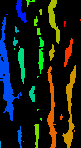}\label{fig_cracks_d}}
\caption{Implementing the method in cracks detecting. (a) is the real cracks image from high-speed photography.
(b) is the segmentation result by a simple hand-tuned thresholding. (c) is the segmentation result by directly implementing our LMW method.
(d) is the segmentation result by improve the LMW method to be an iterative one.}
\label{fig_cracks}
\end{figure}

To test the properties of this new method, we implement it in four different types of
images, as shown respectively in Fig. \ref{fig_Artif}-\ref{fig_cracks}.
The only parameter the user need to preset is the total number of grades $N$,
 then the method operates automatically on the whole image to get the final results.
In Fig. \ref{fig_Artif}-\ref{fig_MRI}, the $N$ is set to be 15 for each. Fig. \ref{fig_cracks}
is a little special, as to be discussed below.

The test on the synthetic image, Fig. \ref{fig_Artif}, mainly searves as a test of
the reasonableness of the method and the correctness of the computer program. As a 
by-product, the local adaptivity of the thresholds is also shown.

To more fully test the local adaptivity of the thresholds, we apply the method
 in the famous grains image, Fig. \ref{fig_Grain}.
Amazingly, all the grains are precisely segmented despite the inhomogeneous illumination.

Then we applied this method to the real breast MRI images, Fig. \ref{fig_MRI},
which is the primary area our segmentation study targeted for. The goal is
to segment out the tumors (usually appear as bright objects on dark background).
Even though the objects are mixed in complex environment, the method is still able to
segment out the different bright objects.
However, plainly implementing the method gets too many false tumors,
and additional information, such as shape and texture, is needed to suppress the false positives.

The implementing of the method on cracks detection is an example of special interest, Fig. \ref{fig_cracks}.
Fig. \ref{fig_cracks_a} is an image from the high-speed photography.
The image reader needs to pick out the cracks in it. However, the cracks are
surrounded by shadows, which smear the edges of the cracks and cause a hard-to-separate background.
Fig. \ref{fig_cracks_b} illustrating the difficulty by a simple thresholding.
Directly implementing our LMW method on the cracks image gets the result Fig. \ref{fig_cracks_c}
(For better comparision, the inside of each contour in Fig. \ref{fig_cracks_c} has been filled by color.
So does Fig. \ref{fig_cracks_d}), which, however, does not overcome the  shadow problem.
Then we modified the method to be iterative: if an object doesn't satisfy a given condition, the object is
further segmented by the LMW method, this goes iteratively until all the objects satisfy the conditions
or a given stop rule is met. Fig. \ref{fig_cracks_d} shows the segmentation result,
which is accepted by the expert image reader, from the iterative version of the method.

\section{Conclusion And Discussion}
\label{sec.end}

A novel segmentation method is proposed in this paper.
It is simple in concept and easy for coding.
As a threshold-based method, its threshold is local adaptive.
It is highly automatic and act on the whole image and can simultaneously segment many objects. The only tunable
parameter is the total grades number $N$ the grayscale range is divided into.
From our experience, $N$ set to be 15 can usually give satisfying result.
The method gives satisfying results in the breast MRI images, though further work
is required to make it more practical. The method has great potential to be
modified to fit more complex tasks, such as cracks detection in high-speed photographs.

This method can be viewed as a basic method, which can serve a much wider area than the one discussed here,
because many image or information processing problems can be turned into the
problem of ``segmentation of bright objects on dark background''. For example,
the detection of architectural distortion in mammography image can be turned into
such a problem after implementing the Gabor filter and phase portrait \cite{Rangayyan:2012}.

This method is also expected to interconnect with other segmentation methods.
For example, we expect the method to mimic a specific active contour model, where the energy
function are set to be the inverse average normal-direction gradient and with the constraint that all the
pixels on the contour belong to the same grayscale grade. Those are the further studies.

\appendices

\ifCLASSOPTIONcompsoc
  \section*{Acknowledgments}
\else
  \section*{Acknowledgment}
\fi

We thank Dr. Guowu Ren for providing us the high-speed photography images and reading the cracks detection results.
This work is support in part by the ...

\ifCLASSOPTIONcaptionsoff
  \newpage
\fi

\end{document}